\documentclass[11pt,a4paper]{article}
\usepackage[hyperref]{acl2020}
\usepackage{times}
\usepackage{latexsym}
\usepackage{mathrsfs}
 \usepackage{bm}
 \usepackage{amssymb}
 \usepackage{booktabs}
 \usepackage{graphicx}

\usepackage{microtype}
\usepackage{verbatim}
\aclfinalcopy 

\title{Knowledge Guided Metric Learning for Few-Shot Text Classification}

\author{
	Dianbo Sui$^{1,2}$,
	Yubo Chen$^1$,
Binjie Mao$^{1,2}$,
Delai Qiu$^{3}$,
    Kang Liu$^{1,2}$,
	Jun Zhao$^{1,2}$\\
	$^1$ National Laboratory of Pattern Recognition, Institute of Automation, \\
	Chinese Academy of Sciences, Beijing, 100190, China \\
	$^2$ University of Chinese Academy of Sciences, Beijing, 100049, China \\
	$^3$ Beijing Unisound Information Technology Co., Ltd, Beijing, 100028, China \\
	\{dianbo.sui, yubo.chen, binjie.mao, kliu,  jzhao\}@nlpr.ia.ac.cn \\
	qiudelai@unisound.com
}
\date{}

\begin{document}
\maketitle
\begin{abstract}
 The training of deep-learning-based text classification models relies heavily on a huge amount of annotation data, which is difficult to obtain. When the labeled data is scarce, models tend to struggle to achieve satisfactory performance. However, human beings can distinguish new categories very efficiently with few examples. This is mainly due to the fact that human beings can leverage knowledge obtained from relevant tasks. Inspired by human intelligence, we propose to introduce external knowledge into few-shot learning to imitate human knowledge. A novel parameter generator network is investigated to this end, which is able to use the external knowledge to generate relation network parameters. Metrics can be transferred among tasks when equipped with these generated parameters, so that similar tasks use similar metrics while different tasks use different metrics. Through experiments, we demonstrate that our method outperforms the state-of-the-art few-shot text classification models.
\end{abstract}

\section{Introduction}
The ability to quickly learn from a small number of examples is a critical feature of human intelligence. This motivates research of few-shot learning \citep{vinyals2016matching, snell2017prototypical, finn2017model, sung2018learning}, which aims to classify unseen data instances (\textit{testing} examples) into a set of new categories with few labeled samples (\textit{support} examples) in each category. In the few-shot setting, the model is trained, when given a specific task, to produce a classifier for that specific task. Therefore, the model is exposed to different tasks during the training phase, and it is evaluated on a non-overlapping set of new tasks.

The key challenge in few-shot learning is to make full use of the limited labeled examples available in the support set to find the ``right'' generalizations as suggested by the task. Metric-based approaches \citep{vinyals2016matching, snell2017prototypical, sung2018learning} are effective ways to address this challenge. In these approaches, examples are represented into a feature space and then predictions are made using a metric between the representations of testing examples and support examples. However, employing metric-based approaches directly in text classification faces a problem that tasks are diverse and  significantly different from each other, since words that are highly informative for one task may not be relevant for other tasks \citep{bao2019distributional}. Therefore, a single metric is insufficient to cope with all these tasks in few-shot text classification \citep{yu2018diverse}.

To adapt metric learning to significantly diverse tasks, we propose a knowledge guided metric learning method. This method is inspired by the fact that human beings approach diverse tasks armed with knowledge obtained from relevant tasks \citep{lake2017building}. We use external knowledge from the knowledge base (KB) to imitate human knowledge, while
the role of external knowledge has been ignored in  previous methods \citep{yu2018diverse, bao2019distributional, geng-etal-2019-induction}. In detail, we resort to distributed representations of the KB instead of symbolic facts, since symbolic facts face the issues of poor generalization and data sparsity. Based on such KB embeddings, we investigate a novel parameter generator network \citep{ha2016hypernetworks, jia2016dynamic} to generate task-relevant relation network parameters. With these generated parameters, the task-relevant relation network is able to apply diverse metrics to diverse tasks and ensure that similar tasks use similar metrics while different tasks use different metrics.

 \begin{figure*}[t]
  	\begin{center} \includegraphics*[clip=true,width=0.95\textwidth,height=0.29\textheight]{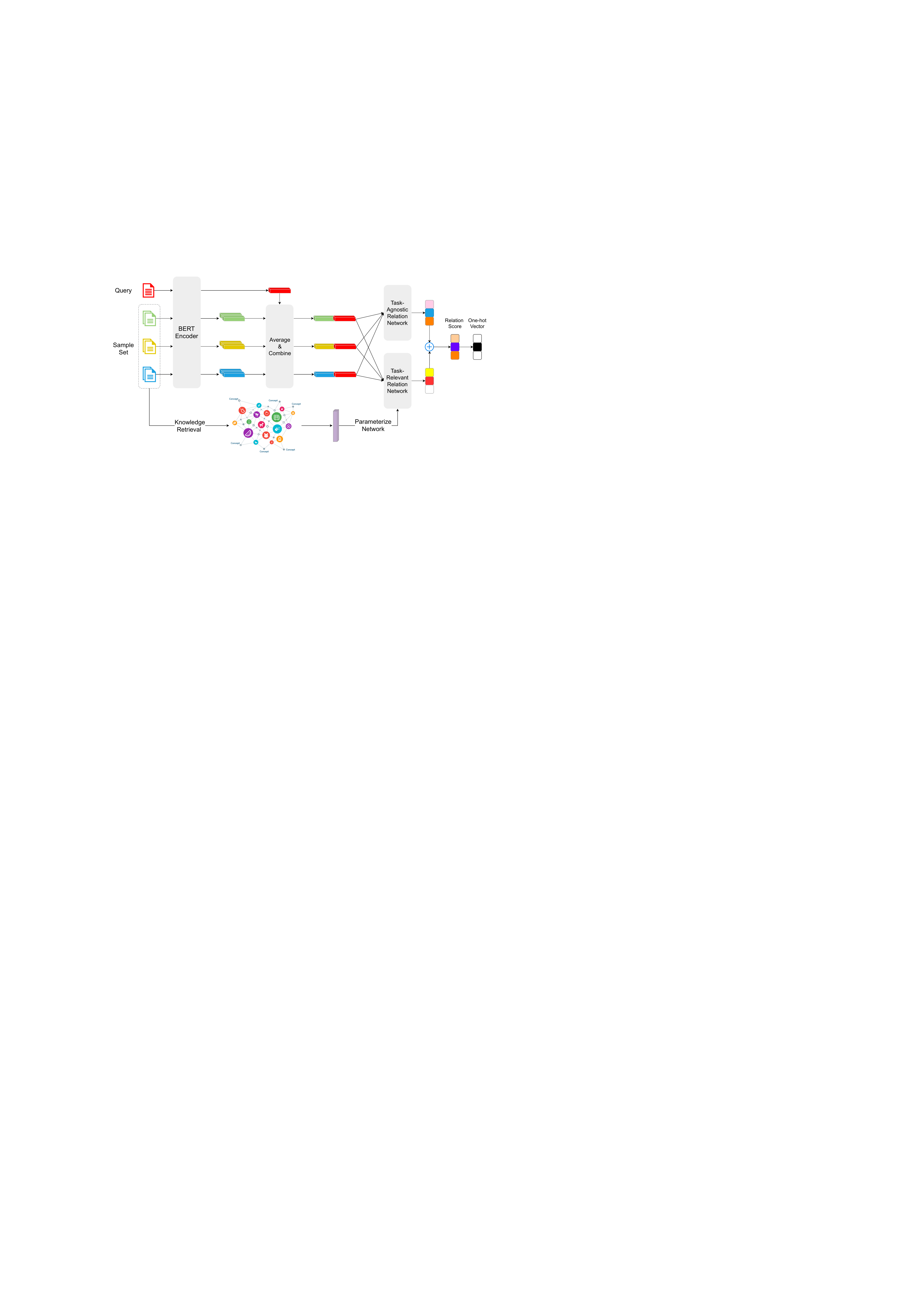}
  	\caption{The main architecture for a C-way N-shot (C = 3, N = 2) problem with one query example} \label{fig}
  \end{center}
  \end{figure*}
In summary, the major contributions of this paper are:
\begin{itemize}
\item Inspired by human intelligence, we present the first approach that introduces external knowledge into few-shot learning.
\item A novel parameter generator network based on external knowledge is proposed to generate diverse metrics for diverse tasks.
\item  The model yields promising results on the \texttt{ARSC} dataset of few-shot text classification.
\end{itemize}






\section{Problem Definition}
Few-shot classification is a task in which a classifier is learned to recognize unseen classes during training with limited labeled examples. Formally, There are three datasets: a training set, a support set, and a testing set. The support set and testing set share the same label space, but the training set has its own label space that is disjoint with support/testing set. If the support set contains N labeled examples for each of C unique classes, the target few-shot problem is called C-way N-shot. In principle, we can train a classifier with the support set only. However, such a classifier usually performs badly on the testing set due to the scarcity of labeled data. Therefore, performing meta-learning on the training set is necessary, which aims to extract transferable knowledge on the training set that will assist the classifier to classify the testing set more successfully.

In meta-learning, testing scenario is simulated during meta-training so the classifier can learn to quickly learn from a few annotations, which is called \textit{episode} based training \citep{vinyals2016matching}. In each training iteration, an episode (or task) is
formed by randomly selecting C classes from the training set with N labelled samples from each of the C classes to serve as the \textit{sample} set $\mathcal{S} = \{(x_i, y_i)\}_{i=1}^{m}$($m=C\times N$), as
well as a fraction of the remainder of those C classes’ samples to act as the \textit{query} set $\mathcal{Q} = \{(x_i, y_i)\}_{i=1}^{n}$, where $x_i$ is a sentence and $y_i \in \{1,...,C\}$ is the corresponding label. This sample/query set split is designed to imitate the support/testing set when testing.
\section{Methodology}
\subsection{Sentence Embedding Network}
  A pre-trained BERT \citep{devlin2019bert} encoder is used to model sentences. Given an input text $x_i=({\rm [CLS]},w_1,w_2,...,w_T,{\rm [SEP]})$ as input, the output of BERT encoder is denoted as $\textbf{H}(x_i) \in \mathbb{R}^{(T+2)\times d_1}$, where $d_1$ is the output dimension of the BERT encoder. We use the first token of the sequence (classification token) as the sentence representation, which is denote as $\textbf{h}(x_i)$.

In meta-learning, each class representation is the mean vector of the embedded sample sentences belonging to its class,
\begin{equation}
\textbf{c}_z = \frac{1}{|\mathcal{S}_z|} \sum \limits_{(x_i,  y_i)\in \mathcal{S}_z} \textbf{h}(x_i) \in \mathbb{R}^{d_1}
\end{equation}
where $\mathcal{S}_z$ denotes the set of examples labeled with class z. Following \citet{sung2018learning}, we use concatenation operator to combine the query sentence representation $\textbf{h}(x_j)$ with the class representation $\textbf{c}_z$.
\begin{equation}
\textbf{p}_{z, j} = {\rm concatenation}(\textbf{c}_z, \textbf{h}(x_j))\in \mathbb{R}^{2d_1}
\label{pair}
\end{equation}

\subsection{Knowledge Guided Relation Network}
This module takes sample set knowledge and combined representation (shown in Equation \ref{pair}) as input, and produces a scalar in range of 0 to 1 representing the similarity between the query sentence and the class representation, which is called relation score. Compared with the original relation network \citep{sung2018learning}, we
decompose the relation network into two parts, 
task-agnostic relation network and task-relevant relation network, in order to serve two purposes. Task agnostic relation network models a basic metric function, while task-relevant relation network adapts to diverse tasks.
\paragraph{Task-Agnostic Relation Network}
The task-agnostic relation network uses a learned unified metric for all tasks, which is the same with the original relation network \citep{sung2018learning}.
With this unified metric, C task-agnostic relation scores $r^{agn}_{z,j}$  are generated for modeling the relation between one query input $x_j$ and the class
representation $c_z$,
\begin{equation}
r^{agn}_{z,j}= RN^{agn}(\textbf{p}_{z,j}|\bm{\theta}^{agn}) \in \mathbb{R}, \quad z = 1,2,...,C
\end{equation}
where $RN^{agn}$ denotes task-agnostic relation network and $\bm{\theta}^{agn}$ are learnable parameters.
\paragraph{Task-Relevant Relation Network}
The task-relevant relation network is able to apply diverse
metrics for diverse tasks armed with external knowledge. In detail, for each sample set, we retrieve a set of potentially relevant KB concepts $K(\mathcal{S})$, where each concept $k_i$  is associated with KB embedding $\textbf{e}_i \in \mathbb{R}^{d_2}$. (we will describe these processes in the following section). We element-wise average over these KB embeddings to form the knowledge representation of this sample set.
\begin{equation}
\textbf{k}_\mathcal{S} = \frac{1}{|K(\mathcal{S})|}\sum \limits_{k_i \in K(\mathcal{S})} \textbf{e}_i \in \mathbb{R}^{d_2}
\end{equation}
Then we use this knowledge representation to generate task-relevant relation network parameters,
\begin{equation}
\bm{\theta}^{rel} = \textbf{M}\cdot\textbf{k}_\mathcal{S}  \in \mathbb{R}^{d_3}
\end{equation}
where $\textbf{M}\in \mathbb{R}^{d_3\times d_2}$ are learnable parameters and $d_3$ denotes the number of parameters of the task-relevant relation network.

With these generated parameters, we use the task-relevant network to generate C task-relevant relation scores $r^{rel}_{z,j}$ for the relation between one query input $x_j$ and the class representation $c_z$,
\begin{equation}
r^{rel}_{z, j} = RN^{rel}(\textbf{p}_{z,j}|\bm{\theta}^{rel}) \in \mathbb{R}, \quad z = 1,2,...,C
\end{equation}
where $RN^{rel}$ denotes task-relevant relation network.

Finally, relation score is defined as:
\begin{equation}
r_{z, j} = Sigmoid(r^{agn}_{z, j} + r^{rel}_{z, j})
\end{equation}
where a sigmoid function is used to keep the score in a reasonable range. Following \citet{sung2018learning}, the network architecture of relation networks is two full-connected layers and mean square error (MSE) loss is used to train the model. The relation score is regressed to the ground truth: the matched pairs have similarity
1 and the mismatched pairs have similarity 0.
\begin{equation}
L = \sum \limits_{z=1} ^ {C} \sum \limits_{j=1} ^ {|\mathcal{Q}|} (r_{z, j}-\textbf{1}(y_j==z))
\end{equation}
\subsection{Knowledge Embedding and Retrieval}
We use NELL \citep{carlson2010toward} as the KB, stored as (subject, relation, object) triples, where each triple is a fact indicating a specific relation between subject and object, e.g., (Intel, competes with, Nvidia). 

\paragraph{Knowledge Embedding} Since symbolic facts suffer from poor generalization and data sparsity, we resort to distributed representation of triples. In detail, given any tripe $(s,r,o)$, vector embeddings of subject $s$, relation $r$ and object $o$ are learned jointly such that the validity of the triple can be measured in the real number space. We adopt the BILINEAR model \citep{yang2015embedding} to measure the validity of triples:
\begin{equation}
f(s,r,o) = \textbf{s}^T diag(\textbf{r})\textbf{o} \in \mathbb{R}
\end{equation}
where $\textbf{s}, \textbf{r}, \textbf{o} \in \mathbb{R}^{d_2}$ are the embeddings associated with $s$, $r$, $o$, respectively, and $diag(\textbf{r})$ is a diagonal matrix with the main diagonal given by the relation embedding $\textbf{r}$. To learn these vector embeddings, a margin-based ranking loss is designed, where triples in the KB are adopted to be positive and negative triplets are constructed by corrupting either subjects or objects.

\paragraph{Knowledge Retrieval} To retrieve knowledge in KB, we first recognize entity mentions from a given passage, link the recognized entity mentions to subjects in KB by
exactly string matching, and then collect the corresponding objects (concepts) as candidates. After this retrieval process, we obtain a set of potentially relevant KB concepts
for each sample set, where each KB concept is associated with a KB embedding.

\section{Experiment}
\subsection{Dataset}
 To make a fair comparison with previous methods, our model is evaluated on widely used \texttt{ARSC} \citep{blitzer2007biographies} dataset. This dataset comprises English reviews for 23 types of products on Amazon. For each product domain, there are three different binary classification tasks. These buckets form 69 tasks in total. Following previous works, we select 12 tasks from four domains (Books, DVDs, Electronics, and Kitchen) as testing set, with only five examples as support set for each label in the testing set. According to meta-training setting, we create 5-shot learning models on the dataset.
 
\subsection{Implementation Details}
In our experiments, we use hugginface's implementation\footnote{https://huggingface.co/pytorch\_transformers/} of BERT (base version) and initialize parameters of the BERT encoding layer with pre-trained models officially released by Google\footnote{https://github.com/google-research/bert}. To represent knowledge in NELL \citep{carlson2010toward}, BILINEAR model \citep{yang2015embedding} is implemented  with the open-source framework OpenKE \citep{han2018openke} to obtain the embedding of entities and relations. The size of embeddings of entities and relations is set to 100. To train our model, We use Adam optimizer \citep{kingma2014adam} with a learning rate of 0.00001.

\subsection{Experiment Results}
\textbf{Baseline}\quad We compare our method to the following baselines: (1) \textbf{Match Network} \citep{vinyals2016matching} is a metric-based attention method for few-shot learning; (2) \textbf{Prototypical Network} \citep{snell2017prototypical} is a deep
metric-based method using sample averages as class prototypes; (3) \textbf{Relation Network} \cite{sung2018learning} is a metric-based method that uses BERT as embedding module and uses two full-connected layers as metric function;(4) \textbf{MAML} \citep{finn2017model} is an optimization-based method through learning to learn with gradients; (5) \textbf{ROBUSTTC-FSL} \citep{yu2018diverse} is an approach that combines adaptive metric methods by clustering the tasks; (6) \textbf{Induction Network} \citep{geng-etal-2019-induction} is a metric-based method by using dynamic routing to learn class-wise representations. (7) \textbf{P-MAML} \cite{zhang2019improving} is the current SOTA method that combine the MAML algorithm with BERT.
\begin{table}[htbp]
\begin{center}
\scalebox{0.8}{
\begin{tabular}{cc}
\toprule
\textbf{Model}        & \textbf{Mean Acc} \\
\hline
Matching Network      & 65.73             \\
Prototypical Network  & 68.15             \\
Relation Network      & 86.09            \\
MAML                  & 78.33             \\
ROBUSTTC-FSL          & 83.12             \\
Induction Network     & 85.63             \\
P-MAML                & 86.65             \\
\textbf{Ours} & \textbf{87.93}     \\
\bottomrule
\end{tabular}}
\caption{Comparison of mean accuracy (\%) on \texttt{ARSC} dataset}
	\label{ARSC}
\end{center}
\end{table}

\noindent \textbf{Analysis}\quad Experiment results on \texttt{ARSC} are presented in Table \ref{ARSC}. We observe that our method achieves the best results amongst all meta-learning models. Compared with P-MAML, our model not only achieve better performance, but also does exempt from requiring backpropagation to update parameters during testing.
Both Induction Network and Relation Network use a single metric to measure the similarity. Compared with these methods, we attribute the improvements of our model to the fact that our model can adapt to diverse tasks with diverse metrics. Compared with ROBUSTTC-FSL, our model leverages knowledge to get implicit task clusters and is trained in an end-to-end manner, which can mitigate error propagation.




\subsection{Ablation and Replacement Studies}
To analyze the contributions and effects of external knowledge in our approach, we perform some ablation and replacement studies, which are shown in Table \ref{ablation}. \textbf{Ablation} means that we delete the task-relevant relation network and the model is reduced to the original relation network. We observe that ablation degrades performance. In order to exclude the factor of reduction in the number of parameters, we conduct a replacement experiment. \textbf{Replacement} means that we replace the task-relevant relation network with a task-agnostic relation network. We find out that increasing the number of parameters can slightly improve performance, but there is still a big gap between our model.
\begin{table}[htbp]
\begin{center}
\scalebox{0.9}{
\begin{tabular}{cc}
\toprule
Model       & Mean Acc \\
\midrule
Ours        & 87.93    \\
Ablation    & 86.09    \\
Replacement & 86.40 \\
\bottomrule
\end{tabular}}
\caption{Ablation and replacement studies of our model on \texttt{ARSC} dataset}
	\label{ablation}
\end{center}
\end{table}

According to the results gained from ablation and replacement experiments, we conclude that the effectiveness of our model is credited with introducing external knowledge rather than increasing the number of model parameters.

\section{Conclusion}
Inspired by human intelligence, we introduce external knowledge into few-shot learning. A parameter generator network is investigated to this end, which can use external knowledge to generate relation network parameters. With these parameters, the relation network can handle diverse tasks with diverse metric. Through various experiments, we demonstrate that our model outperforms the current SOTA few-shot text classification models.

\newpage
\bibliography{anthology,acl2020}
\bibliographystyle{acl_natbib}

\end{document}